\documentclass[10pt,twocolumn]{article}

\usepackage{cvpr}
\usepackage{times}
\usepackage{epsfig}
\usepackage{graphicx}
\usepackage{amsmath}
\usepackage{amssymb}

\usepackage{subfigure}
\usepackage{multirow}
\usepackage{tabularx}
\usepackage{times}
\usepackage[misc]{ifsym}

\usepackage[breaklinks=true,bookmarks=false]{hyperref}

\cvprfinalcopy 

\def\cvprPaperID{****} 

\begin{document}

\title{Grand Challenge of 106-Point Facial Landmark Localization}

\author{Yinglu Liu$^{1}$, Hao Shen$^{1}$, Yue Si$^{1}$, Xiaobo Wang$^{1}$, Xiangyu Zhu$^{2}$, Hailin Shi$^{1}${\Letter} , \\
	Zhibin Hong$^{3}$, Hanqi Guo$^{3}$, Ziyuan Guo$^{3}$, Yanqin Chen$^{3}$, Bi Li$^{3}$, Teng Xi$^{3}$, \\
	Jun Yu$^{4}$, Haonian Xie$^{4}$, Guochen Xie$^{4}$, Mengyan Li$^{4}$, Qing Lu$^{5}$, Zengfu Wang$^{4}$,\\
	Shenqi Lai$^{6}$, Zhenhua Chai$^{6}$, Xiaoming Wei$^{6}$.\\ \\
	$^{1}$JD AI.
	$^{2}$Institute of Automation, Chinese Academic of Sciences.\\
	$^{3}$Department of Computer Vision Technology (VIS), Baidu Inc.\\
	$^{4}$University of Science and Technology of China.
	$^{5}$IFLYTEK CO.,LTD.\\
	$^{6}$Vision and Image Center of Meituan.}

\maketitle

\begin{abstract}
Facial landmark localization is a very crucial step in numerous face related applications, such as face recognition, facial pose estimation, face image synthesis, \textit{etc}. However, previous competitions on facial landmark localization (\textit{i.e.}, the 300-W, 300-VW and Menpo challenges) aim to predict 68-point landmarks, which are incompetent to depict the structure of facial components.
In order to overcome this problem, we construct a challenging dataset, named JD-landmark. Each image is manually annotated with 106-point landmarks. This dataset covers large variations on pose and expression, which brings a lot of difficulties to predict accurate landmarks. We hold a 106-point facial landmark localization competition\footnote{https://facial-landmarks-localization-challenge.github.io/} on this dataset in conjunction with IEEE International Conference on Multimedia and Expo (ICME) 2019. The purpose of this competition is to discover effective and robust facial landmark localization approaches.
\end{abstract}

\section{Introduction}
Facial landmark localization, which is to predict the coordinates of a set of pre-defined key points on human face, plays an important role in numerous facial applications. For example, it is commonly used for face geometric normalization which is a crucial step for face recognition. Besides, landmarks are often employed to support more and more interesting applications due to their abundant geometric information, \textit{e.g.}, 3D face reconstruction and face image synthesis.
In recent years, the deep learning methods have been largely developed and the performances are continuously improved in facial landmark localization task. However, facial features vary greatly from one individual to another. Even for a single individual, there is a large amount of variations due to the pose, expression, and illumination conditions. There still exist many challenges to be addressed.
The iBUG group\footnote{https://ibug.doc.ic.ac.uk/} held several competitions on facial landmark localization.
Nevertheless, they all focus on the 68-point landmarks which are incompetent to depict the structure of facial components, \textit{e.g.}, there is no points defined on the lower boundary of eyebrow and the wing of nose. To overcome the above problems, we construct a challenging dataset and hold a competition of 106-point facial landmark localization in conjunction with ICME 2019 on this dataset.
The purpose of this competition is to promote the development of research on 106-point facial landmark localization, especially dealing with the complex situations, and discover effective and robust approaches in this field.
It has attracted wide attention from both academia and industry. Finally, more than 20 teams participated in this competition. We will introduce the approaches and results of the top three teams in this paper.

\section{JD-landmark Dataset}
\begin{figure}[t]
	\centering
	\subfigure[examples of training set]{
		\includegraphics[width=0.45\linewidth]{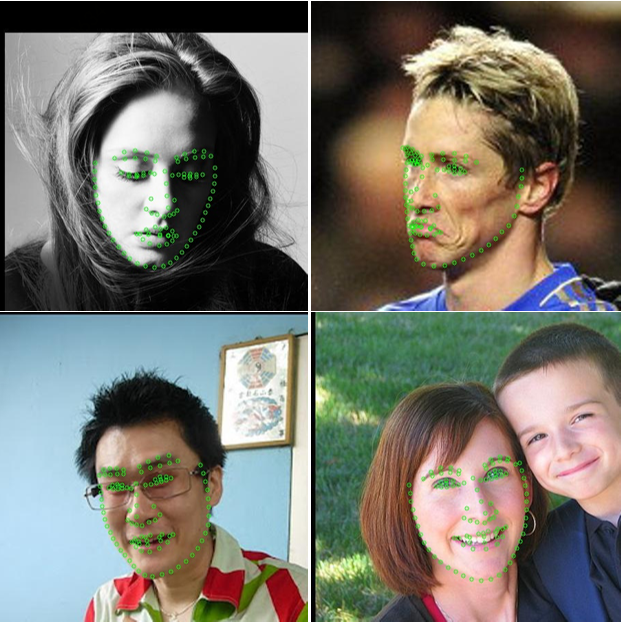}\label{fig:train}
	}
	\subfigure[examples of validation set]{
		\includegraphics[width=0.45\linewidth]{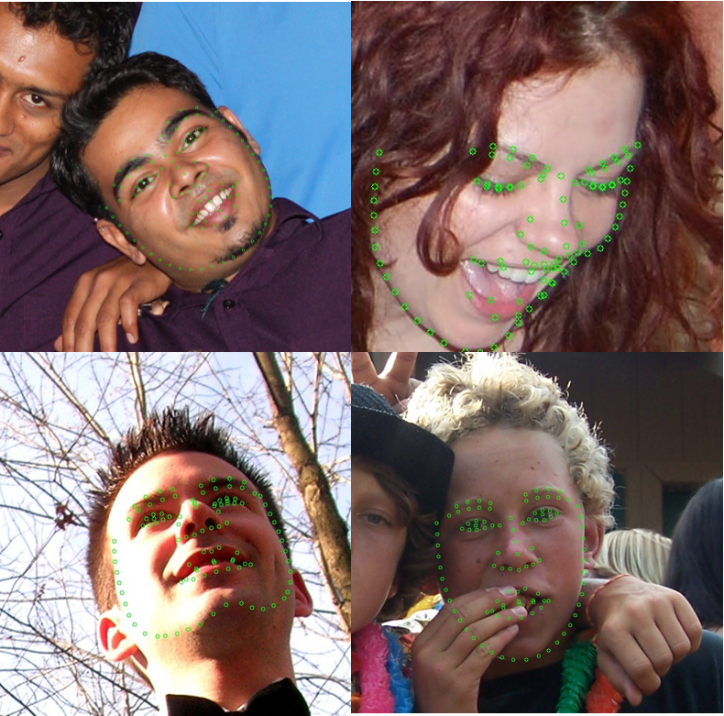}\label{fig:test}
	}
	\caption{examples of JD-landmark dataset.}
	\label{fig:performance_hg}
\end{figure}

\begin{figure}[t]
	\centering
	\includegraphics[width=0.7\linewidth]{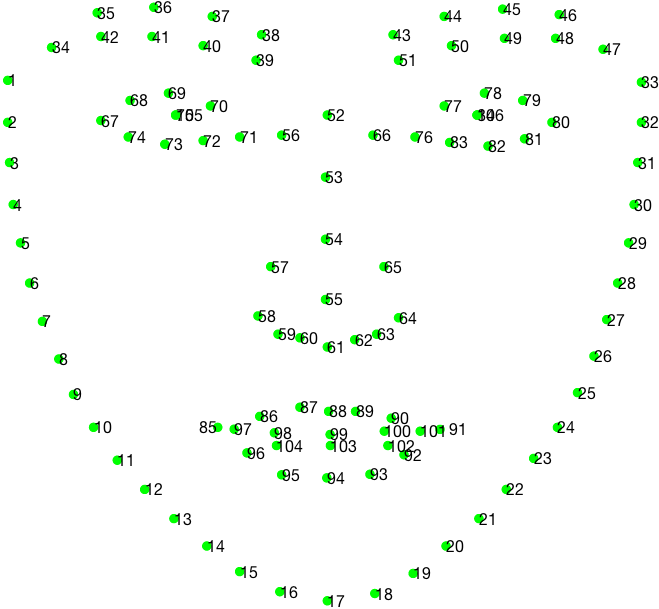}
	\caption{The 106-point landmark make-up.} \label{fig:106p}
\end{figure}

In order to develop advanced approaches for dense landmark localization, we construct a new dataset, named JD-landmark\footnote{https://sites.google.com/view/hailin-shi}. It consists of about 16,000 images. As Tab.~\ref{tab:pose} shows, our dataset covers large variations of pose, in particular, the percent of images with pose angle large than $30^o$ is more than 16\%.
The training, validation and test sets are described as follows:

\begin{itemize}
	\item \textbf{Training set:} We collect an incremental dataset based on 300W~\cite{sagonas2013300,sagonas2016300,zhu2016face}, composed of LFPW~\cite{belhumeur2013localizing}, AFW~\cite{ramanan2012face}, HELEN~\cite{le2012interactive} and IBUG~\cite{sagonas2013semi}, and re-annotate them with the 106-point mark-up as Fig.~\ref{fig:106p} shows. This dataset, containing 11,393 face images, is applied as the training set. It is accessible to the participants (with landmark annotations). Fig.\ref{fig:train} shows some examples of training set.
	
	\item \textbf{Validation set:} 2,000 web face images, covering large variations of pose, expression and occlusion, are selected from open source web face database~\cite{kemelmacher2016megaface}. The participants could optimize the parameters on this set before the final evaluation. Fig.~\ref{fig:test} shows some examples of validation set.
	
	\item \textbf{Test set:} It contains 2,000 web face images as well, which is blind to participants throughout the competition. It will be used for the final evaluation.
\end{itemize}
We emphasize that we provide the bounding boxes obtained by our detector for training/validation/test sets. However, participants have the choice of employing other face detectors.

\begin{table}[t]
	\centering
	\caption{Statistics on pose variations.}
	\label{tab:pose}
	\begin{tabular}{ccc}
		\hline
		0$^o\sim15^o$ & 15$^o\sim30^o$ & $>30^o$ \\
		\hline
		34.9\%&48.7\%&16.4\% \\
		\hline
	\end{tabular}
\end{table}

\section{Evaluation Results}

\subsection{Evaluation criterion} \label{sec:criterion}
All submissions are assessed on the total 106-point landmarks as Fig.~\ref{fig:106p} shows. The average Euclidean point-to-point error normalized by the bounding box size is taken as the metric, which is computed as:
\begin{align}
	NME = \frac{1}{N} \sum_{k=1}^N \frac{\|y_k-\hat{y}_k\|_2}{d} \label{eq:nme}
\end{align}
where $k$ refers to the index of landmarks. $y$ and $\hat{y}$ denotes the ground truths and the predictions of landmarks for a given face image. In order to alleviate the bias for profile faces caused by the small interocular distance, we employ the square-root of the ground truth bounding box as the normalization factor $d$, computed as $d = \sqrt{w_{bbox} \times h_{bbox}}$. Here $w_{bbox}$ and $h_{bbox}$ are the width and height of the enclosing rectangle of the ground truth landmarks. If no face is detected, the NME will be set to infinite.
The Cumulative Error Distribution (CED) curve corresponding to the percentage of test images of which the error is less than 8\% is produced, and the Area-Under-the-Curve (AUC) from the CED curve is calculated as the final evaluation criterion. Besides, further statistics from the CED curves such as the failure rate and average NME are also presented for reference.

\begin{table*}[htbp]
	\centering
	\caption{Final results for the 106-point Facial Landmark Localization competition. The top three teams are listed according to their rank in the challenge. Methods are ranked according to the AUC of the CED curve. The Failure rate and NME are also presented for reference.}
	\label{tab:results}
	\begin{tabular}{ccccc}
		\hline
		Rank & Participant & AUC(\%) & Failure rate(\%) & NME(\%) \\
		\hline
		\multirow{2}{*}{1} & Z. Hong, Z. Guo, Y. Chen, H. Guo, B. Li and T. Xi & \multirow{2}{*}{84.01} & \multirow{2}{*}{0.10} & \multirow{2}{*}{1.31} \\
		&Department of Computer Vision Technology (VIS), Baidu Inc.&&& \\
		\hline
		\multirow{2}{*}{2} & J. Yu, H. Xie, G. Xie, M. Li, Q. Lu and Z. Wang & \multirow{2}{*}{82.68} & \multirow{2}{*}{0.05} & \multirow{2}{*}{1.41} \\
		&University of Science and Technology of China.&&& \\
		\hline
		\multirow{2}{*}{3} & S. Lai, Z. Chai and X. Wei & \multirow{2}{*}{82.22} & \multirow{2}{*}{0.00} & \multirow{2}{*}{1.42} \\
		&Vision and Image Center of Meituan&&& \\
		\hline
	\end{tabular}
\end{table*}

\subsection{Participation}
A total of 23 teams participated in this challenge. Due to the space limitation, we will briefly describe the submitted methods of the top three winners in this subsection.

Hong~\textit{et al.} proposed a Multi-Stack Face Alignment method based on autoML~\cite{he2018amc}. It consists of several stacked hourglass models~\cite{newell2016stacked,yang2017stacked,bulat2017far} and performs landmark localization from coarse to fine. The final results are obtained by fusing the outputs based on a voting strategy which could find the most confident cluster and reject outliers. The base models are developed with the help of autoML and trained with a well-designed data augmentation scheme. In addition, one of the base models is jointly trained with segmentation as multi-task learning to take advantage of extra supervision. Equipped with the above designs, the method could perform precise facial landmark localization in various conditions including those with large pose and occlusions.

\begin{figure}[htbp]
	\centering
	\includegraphics[width=0.9\linewidth]{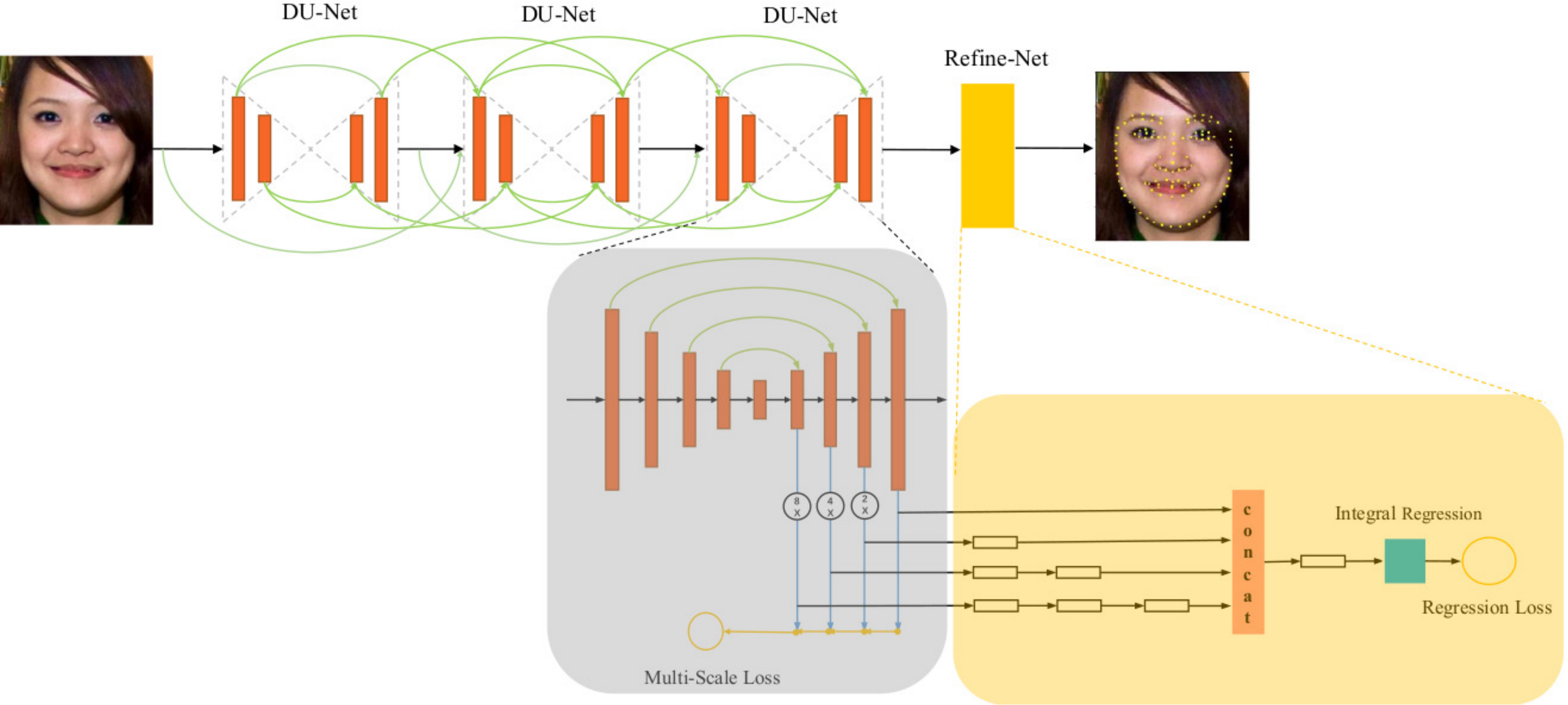}
	\caption{Densely U-Nets Refine Network.} \label{fig:yu}
\end{figure}

Yu~\textit{et al.} employed a Densely U-Nets Refine Network (DURN) for facial landmark localization. As shown in Fig.~\ref{fig:yu}, it involves two sub-networks: DU-Net and Refine-Net. The DU-Net is based on Tang \textit{et al.}~\cite{tang2018quantized}, where the original intermediate supervision is modified to multi-scale intermediate supervision. It means that each DU-Net employs four intermediate supervision rather than one. The Refine-Net is based on Chen \textit{et al.}~\cite{chen2018cascaded}, and Yu \textit{et al.} add the integral regression~\cite{sun2018integral} after the Refine-Net to obtain the keypoint coordinates instead of the heatmap via argmax function. In addition, the regression loss is computed by the coordinate rather than the heatmap. Finally, Yu \textit{et al.} ensemble 7 models with the similar structure.

\begin{figure}[htbp]
	\centering
	\includegraphics[width=0.8\linewidth]{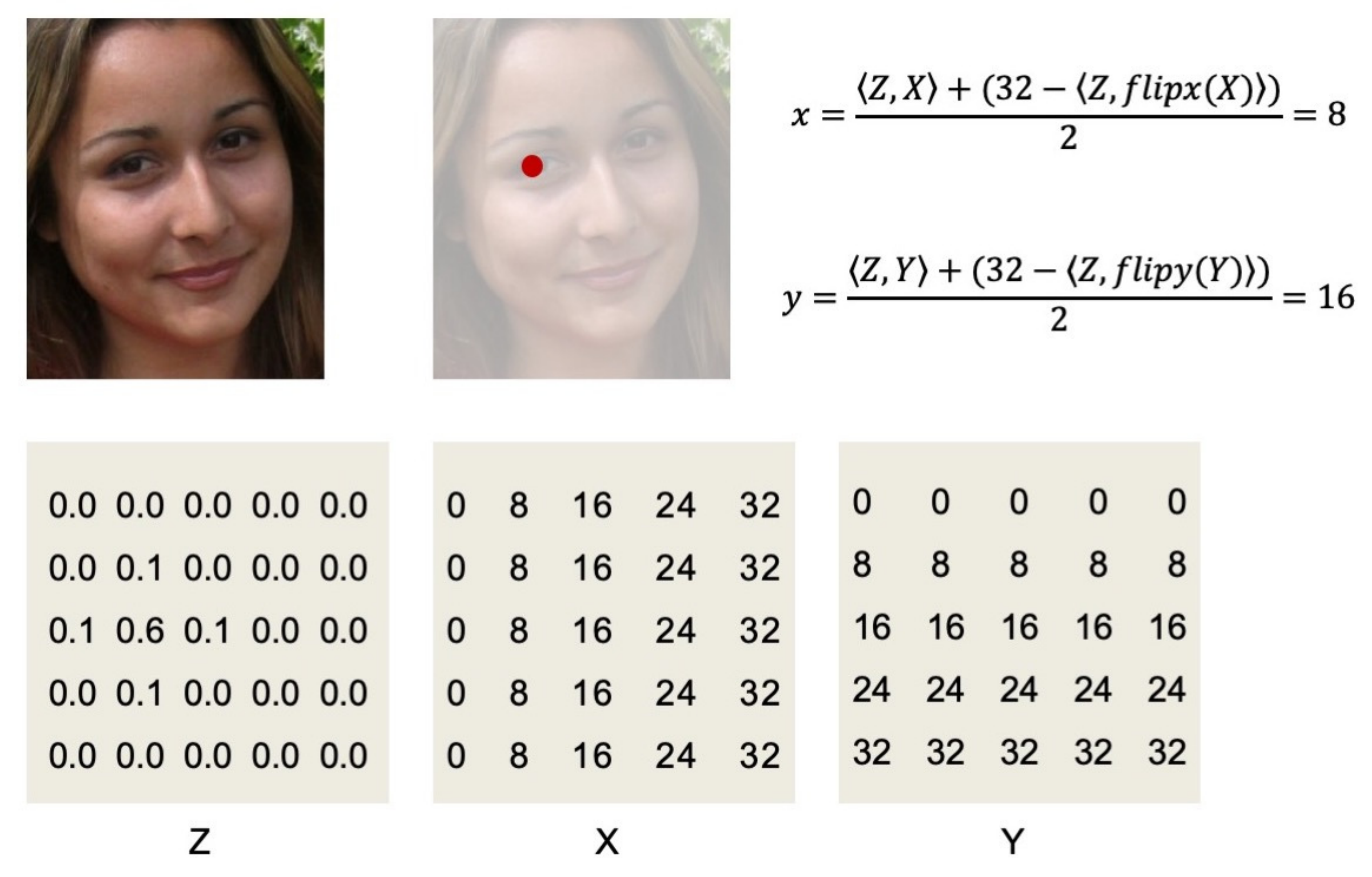}
	\caption{Flipx means flipping matrix horizontally and flipy means flip matric vertically.} \label{fig:lai}
\end{figure}

Lai \textit{et al.} proposed an end-to-end trainable facial landmark localization framework, which has achieved promising localization accuracy even under challenging wild environments  (\textit{e.g.} unconstrained pose, expression, lighting and occlusion).  Different from the classical four stage stacked HGs \cite{newell2016stacked}, they propose to use the hierarchical module~\cite{bulat2017binarized} rather than the standard residual block, which will generate the probability heatmap for each landmark and can make the non-linearity stronger.  Besides, in previous work~\cite{bulat2017binarized} researchers use argmax and post-process operations (e.g. rescale) to get final results, which may decrease the performance by the coordinate quantization. In order to overcome this problem, dual soft argmax function is proposed to map probability of heatmap to numerical coordinates, which is shown in Fig.~\ref{fig:lai}. For a gaussian response in an image, with matrix X and matrix Y,  the coordinates x and y can be computed directly.  Finally, three models (\textit{i.e.} $64 \times 64$ SA, $64 \times 64$ DSA and $128 \times 128$ DSA) are trained in total, where the number means the size of output heatmaps, SA means soft argmax and DSA means dual soft argmax. The weighted predictions of three models will be used as the final results.

\subsection{Results}
\begin{figure}[htbp]
	\centering
	\includegraphics[width=0.85\linewidth]{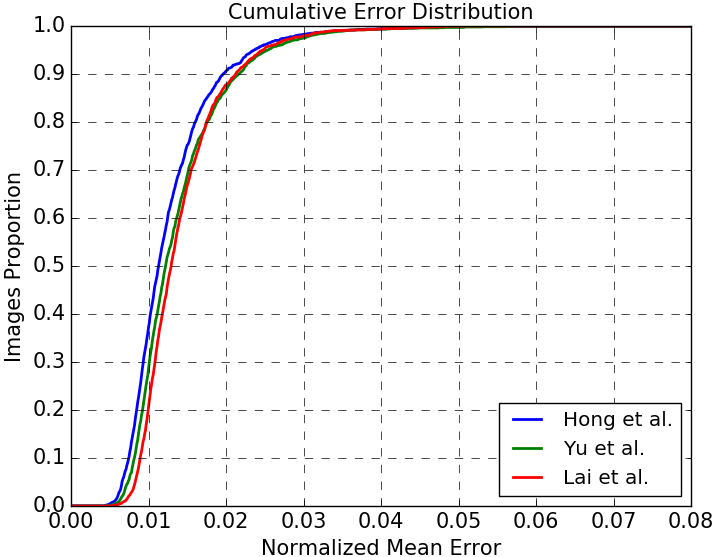}
	\caption{The Cumulative Errors Distribution curves of top three teams on the JD-landmark test set.} \label{fig:ced}
\end{figure}

As is mentioned in Sec.~\ref{sec:criterion}, the submissions are ranked according to the AUC of the CED curve with the threshold of 8\%. The winner is Hong \textit{et al.} from Baidu Inc. Second place goes to Yu \textit{et al.} from University of Science and Technology of China. Lai \textit{et al.} from Meituan achieves the 3rd place.
Fig.~\ref{fig:ced} draws the CED curves of the top three teams on the JD-landmark test set. In order to comprehensively evaluate the submissions, we also report the average NME defined as Eq.~(\ref{eq:nme}) and the failure rate (if the average NME is larger than 8\%, the picture will be taken as a failure prediction.) in Tab.~\ref{tab:results}. We can see that Hong \textit{et al.} achieved the highest AUC of 84.01\%, higher than Yu \textit{et al.} and Lai \textit{et al.} by 1.33\% and 1.79\%, respectively. Hong \textit{et al.} also performed the best on NME (1.31\%), lower than Yu \textit{et al.} and Lai \textit{et al.} by 0.1\% and 0.11\%, respectively.

\section{Conclusion}
In this paper, we summarize the grand challenge of 106-point facial landmark localization in conjunction with ICME 2019. We construct and release a new facial landmark dataset, named JD-landmark. Compared with previous challenges on facial landmark localization, our work pays attention on 106-point landmarks which contain more structure information than the 68-point landmarks. Meanwhile, our dataset covers large variations of poses and expressions, which bring a lot of difficulties for participants. Finally, more than 20 teams submitted their binaries or models. We introduced the methods together with the performance of top three teams in this paper. We hope this work could make contributions on the development of facial landmark localization.

{\small
\bibliographystyle{ieee_fullname}
\bibliography{egbib}
}

\end{document}